# Select-and-Combine (SAC): A Novel Multi-Stereo Depth Fusion Algorithm for Point Cloud Generation via Efficient Local Markov Netlets

Mostafa Elhashash and Rongjun Qin, *Senior Member, IEEE*

*Abstract*—: **Many practical systems for image-based surface reconstruction employ a stereo/multi-stereo paradigm, due to its ability to scale for large scenes and its ease of implementation for out-of-core operations. In this process, multiple and abundant depth maps from stereo matching must be combined and fused into a single, consistent, and clean point cloud. However, the noises and outliers caused by stereo matching and the heterogenous geometric errors of the poses present a challenge for existing fusion algorithms, since they mostly assume Gaussian errors and predict fused results based on data from local spatial neighborhoods, which may inherit uncertainties from multiple depths resulting in lowered accuracy. In this paper, we propose a novel depth fusion paradigm, that instead of numerically fusing points from multiple depth maps, selects the best depth map per point, and combines them into a single and clean point cloud. This paradigm, called select-and-combine (SAC), is achieved through modeling the point level fusion using local Markov Netlets, a micro-network over point across neighboring views for depth/view selection, followed by a Netlets collapse process for point combination. The Markov Netlets are optimized such that they can inherently leverage spatial consistencies among depth maps of neighboring views, thus they can address errors beyond Gaussian ones. Our experiment results show that our approach outperforms existing depth fusion approaches by increasing the F1 score that considers both accuracy and completeness by 2.07% compared to the best existing method. Finally, our approach generates clearer point clouds that are 18% less redundant while with a higher accuracy before fusion.**

## I. INTRODUCTION

Image-based 3D reconstruction is a fundamental yet challenging problem in photogrammetry and computer vision. Multi-view stereo (MVS) algorithms aim to accurately recover 3D dense representations from redundant image observations. The recent advances in deep learning and

*Corresponding author: Rongjun Qin*

Mostafa Elhashash is with Geospatial Data Analytics lab, and Department of Electrical and Computer Engineering. The Ohio state University, Columbus, USA.

Rongjun Qin is with the Geospatial Data Analytics Lab, Department of Civil, Environment and Geodetic Engineering, Department of Electrical and Computer Engineering, and Translational Data Analytics Institute, The Ohio state University, Columbus, USA (email: qin.324@osu.edu ) the increasingly available benchmark datasets have pushed forth the theoretically attainable accuracy of the MVS algorithms to another level [1-3]. In order to practically process large-format and mega-pixel images, the existing system favors a depth fusion approach for scalability of computation and ease of implementation for out-of-core processing. In this approach, depth maps of each view are generated through stereo matching or MVS algorithms (using a small number of images), and then are fused and combined into a single point cloud that is desired to be consistent and clean. However, this is a non-trivial task since these depth maps are independently generated and are subject to errors beyond typical Gaussian noises. For example, these depth maps can be biased due to systematic errors of the poses, can be associated with variable uncertainties due to differences in stereo configurations (convergency angles) which is a common case in aerial photogrammetry, and be error-prone due to scene complexities. Existing methods for depth fusion tend to oversimplify these errors to be zero-mean and follow Gaussian distributions. As a result, the prevalent approaches in the depth fusion step of the MVS 3D reconstruction, such as using simple median [4, 5], oriented filtering [6], TSDF (Truncated Signed Distance Function) fusion [7, 8], etc., fail to effectively accommodate systematic biases and large blunders. Oftentimes this leads to observable uncertainty of the fused point clouds (an example is shown in Fig. 1), which consequently leads to lowered absolute/relative accuracy of both the point clouds and the resulting mesh. Fig. 1 shows that sometimes the depth map (red) of an individual view, may be much more consistent with the ground truth (green), than a fused point cloud (blue) from multiple depth maps. Certain depth maps from single views can be optimal for certain reasons, partially due to their more optimal local camera networks, more consistent radiometric quality, and more importantly, free from possible systematic biases introduced by the depth of other views. Thus, instead of fusing depth maps from various views, selecting the optimal depth maps to seed the fusion may best leverage the quality of these optimal depth maps in the fusion result.

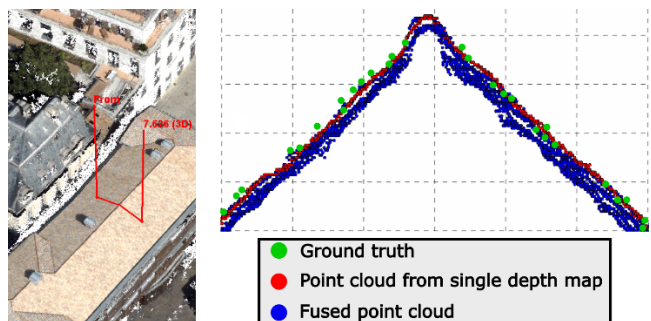

Fig. 1. Illustration of inconsistencies of the fused point clouds from different depth maps. The right column shows a profile of a cross-section of the rooftop on the left. We observe that the generated point cloud from a single depth map may be consistent with the ground truth compared to the point cloud fused from different depth maps.

Inspired by this, we hence propose a novel paradigm for MVS depth fusion, called Select-and-Combine (SAC), that instead of numerically fusing every single value of the redundant depth maps, selects the optimal set of depth maps seedling a consistent, accurate, and clean point cloud. The selected depth maps should minimize the overlapped region to avoid point/pixel level fusion and should minimize the systematic biases among multiple depth maps. Therefore, we formulate SAC as an optimization problem: for each point in the space, we select its values from the optimal depth map (considered as a label) observing it, such that the selection fulfills a few criteria: first, it yields the smallest number of depth maps in the overall set to minimize depth uncertainties; second, spatially adjacent depth map candidates should be minimally biased in their overlapping region. This can be formulated as lightweight Markovian Networks [9], called Netlets, where potentially redundant points in local neighborhoods are modeled through self-dependences that favor a homogenous label (depth map), as well as minimized biases. The solution of the Markovian Netlets associates each point candidate with a selected view, and the fusion of these points is achieved through a Netlet collapse process that assigns the fused points of these Netlets from the minimally selected views. As a result, the fused points are inherently associated with the minimal number of individual depth maps to achieve minimized uncertainty (Fig. 1). Since there exist millions of Markovian Netlets in this process, to accelerate it for practical aerial mapping applications, we propose a fast traversing scheme through the complex 3D space, utilizing spatial clusters assisted through inter-depth map reprojections to achieve linear speed. Our proposed SAC approach provides a new paradigm for MVS fusion with greater extendibility and less tunable parameters. For example, since the fusion problem is reduced to a labeling problem that selects the best, fewer adaptations are needed to numerically fuse multiple depths of different quality. Moreover, modeling the problem as a Markovian network would allow flexibility to incorporate learnable marginal distributions of the network and the inference functions when available datasets are present. We validate our approach by comparing it against typical MVS depth fusion methods through extensive experiments, and both qualitative and quantitative results show that it outperforms existing methods in yielding consistent, accurate, and clean point clouds.

The rest of this paper is organized as follows: Section II reviews related works in MVS and depth fusion techniques; Section III introduces our proposed SAC approach for MVS depth fusion in detail; Section IV presents the experimental results, comparative studies, and analysis using various datasets. Finally, Section V concludes this work by analyzing the pros and cons, as well as drawing future works.

## II. RELATED WORK

Thanks to the recent development in the Structure-from-Motion/photogrammetry and SLAM (Simultaneous Localization and Mapping) pipelines, there have been consistent contributions to each component of these paradigms. Depth fusion methods in MVS reconstruction and SLAM are most relevant to our work, which can be generally categorized into three main approaches: depth filtering, fusion through geometric consistency check, and TSDF methods.

The first two approaches are mostly used in MVS frameworks. Typically, the majority of traditional MVS approaches either rely on semi-global matching (SGM) as the core matching approach [10, 11], or PatchMatch-based approaches such as [12-16]. The typical pipeline involves processing multiple stereo pairs that yield a depth map for each pair. This has recently been progressed through deep learning approaches, which rely on convolutional layers to extract deep features and build cost volumes for depth map generation [17-20]. Typically, these individually generated depth maps will contribute to an optimal point cloud through depth fusion, which is critical yet challenging, as these depth maps are subject to various levels of uncertainties and inconsistencies [6].

**Depth fusion through filtering.** Filtering-based approaches determine the fused depth value as a function of its local neighborhood across multiple depth images. [6] introduced a fusion approach based on median filtering. They used a tree-based structure to store the points and filter points in a cylinder along the line of sight or surface normal. [4] adopted a similar approach while classifying disparity maps to different quality levels based on total variation through a weighted median approach. Although median filtering is robust to outliers, it cannot handle measurements from different distributions and is inefficient in eliminating inconsistencies caused by biases in the estimated camera poses from bundle adjustment.

**Depth fusion through geometric consistency check.** The geometric consistency check approaches fuse 3D points from individual depth maps that comply with a certain rule, e.g. spatial proximity. [15] adopted a fusion strategy for depth maps by projecting the estimated depth value from one image to its neighbors and rejecting points that are occluded or are very close in values assuming they are redundant. [13] applied a fusion strategy to merge redundant points by identifying whether they are consistent or not based on angles between normal and disparity values. Then, consistent points are averaged in the object space for consistent views to reduce noise. Similarly, [12, 21] identified consistent points using the relative depth difference and angles between normal and reprojection errors. The aforementioned strategies are often adopted by several MVS approaches [16, 18, 22] with some modifications such as only relying on disparity values and neglecting the normal or changing the formulation of geometric consistency check. Although these fusion approaches produce reasonable results, they are sensitive to the choice of a few thresholds that have a high impact on the quality of the point clouds. For example, setting a lower threshold for relative depth or disparity errors can affect completeness whereas higher thresholds yield inaccurate point cloud. Therefore, these values have to be tuned differently according to the scale of the dataset.

**TSDF-based fusion approaches.** TSDF (Truncated Signed Distance Function) is an implicit representation of the scene and was favored in the robotics community for rapid depth fusion over high-frequency depth frames [7, 8, 23, 24]. It models the scene with a continuous function over the full 3D space, where the equilibrium points (function values equal to zero) represent the surface. In comparison to other per-point approaches, the fusion can be performed through simple algorithmic operations on the function fields, which are fundamentally more efficient to determine the fused TSDF representations. However, because it mostly uses simple algorithmic operations on the function field (e.g. mean/median etc.), it requires a large number of images to eliminate noises [25] and can be prone to outliers and problematic for reconstructing thin geometry [26-28]. Therefore, it is mostly suitable for continuous scans where measurements can be modeled using a Gaussian distribution, while for depth fusion using discrete depth maps from sparse views, it does not possess the advantage to use more intricate fusion to achieve higher accuracy for measurements with non-Gaussian errors.

**Our proposed approach:** Most of the existing approaches mentioned above assume an averaging/median process over the measurements, which may fall short for depth fusion with sparse observations and cannot address the systematic uncertainty of MVS depth maps (Fig. 1). In contrast to fusing multiple depth values, our proposed approach adaptively selects the best one over the few, which can bypass the "averaging/median" process. This by concept, can be more robust to handle measurements that are either not following Gaussian, or are too sparse to determine by "averaging/median".

## III. METHODOLOGY

Our SAC (Select-and-Combine) approach takes a set of oriented images (i.e., images with estimated camera poses) and their corresponding depth maps produced from typical MVS approaches as inputs, producing a unified point cloud with minimized redundancy. Considering a 3D point $\mathbf{p}$ from a depth image $\boldsymbol{d}_i$, it can be visible from different depth maps. The central task of our approach is to decide a depth (or a view) that this point $\mathbf{p}$ should be re-associated with, such that the depth maps hosting these 3D points are as minimal and consistent as possible. As a result, 3D points from these minimal set of views share the consistency of the depth in the image space (as shown in Fig. 2). This view re-association problem can be constructed per point within a micro-group of networked points from different views but with high spatial proximity (Fig. 2), thus decisions can be made within this small group of points upon which a minimal set of views best covers this group of points. We define these networked points as a Netlet, which can be either cyclic or acyclic. To assign a view for each point, this Netlet can be best modeled as a Markovian dependence (thus called, Markov Netlet), where spatially closed points, despite originally coming from different views, may be assigned the same view to achieve a view consistency.

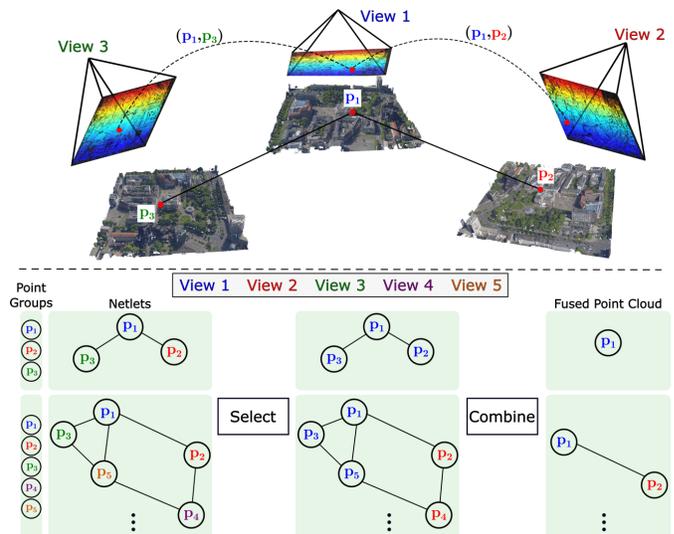

**Fig. 2.** The workflow of the proposed SAC approach for depth fusion. The inputs to our approach are oriented views and their corresponding depth maps are shown at the top. Edges are built between nodes based on adaptive spatial connectivity between corresponding points in 3D space to formulate a point group. $(\mathbf{p}_1 - \mathbf{p}_5)$ are 3D points represented as nodes, each colored differently according to the assigned label (depth map/view). At the bottom, we show the main steps of our SAC approach for two Markov Netlets as an example formulated using two point groups. Initially, each node is assigned to a label based on its corresponding view. After the optimization, a label is selected for each node to re-associate it with a new view. Finally, nodes with the same label are combined into a single node to formulate the final fused point cloud.

Our SAC approach formulates these local networks as Markov Netlets that can efficiently embed the direct interactions between points to ensure the spatial smoothness of the solution. In this formulation, 3D points are regarded as nodes, and the spatial information between the 3D points is represented as edges. The formulation of these Netlets enables us to *select* a label (depth map/view) for each node in the network, and the solution of this labeling problem will be used to guide a "node collapse" process that unifies the same-labeled points within a network into a single one and *combine* the selected points into a unified point cloud. These locally selected points overall contribute to the fusion of redundant points in the space, and because the points are locally selected from a limited number of views, they achieve a better point cloud consistency (effects of this explained in Fig. 1).

In this section, we briefly describe a typical pre-processing step generating per-view depth maps in Section III.A, and then in Section III.B, we present the details proposed SAC approach and the formulation of the Markov Netlets for depth fusion. Finally, Section III.C describes how we extend our SAC approach to work for large-scale data by adapting the process on superpixel clusters.

### A. Depth Estimation

We use Census-based [29] Semi-global matching [11] to generate pairwise disparity and depth maps. Since it requires large memory for computations, we adopt a hierarchical approach as described in [10, 30]. Moreover, the per-view depth estimation may utilize other approaches such as [16, 20, 22]. For depth map computation, each image in the input data is regarded as a master view and matched to a set of neighboring images based on the exterior orientation parameters. This will yield a number of stereo-based depth maps per view. To obtain a higher-quality per-view depth map as input to our SAC approach, we apply a two-step outlier removal process. Firstly, we performed median filtering over these stereo-based depth maps. It should be noted that this step is not necessary if only one neighbor per view is used, which may oftentimes be sufficient [15]. Secondly, for all experiments conducted in this work, each view explores seven neighboring views to produce triangulated points with at least three rays.

### B. Markov Netlets for Depth Map Fusion

As mentioned before, these depth maps are highly redundant in 3D. In order to identify and fuse identical points, we establish local networks called Netlets for 3D point groups. A point group is defined as a set of 3D points that are potentially identical or highly redundant. To find such point groups, we start with each point from each depth map and then find their corresponding points across different depth maps (see Fig. 2). Points within a point group will be connected as a Markovian Netlet, where edges will be established between each pair of points (Fig. 2). A few strategies can be adopted to find corresponding points among different images and build these edges; for instance, one can utilize the generated disparity maps from the dense stereo matching to obtain correspondences of pixels referencing depth points among neighboring views. In addition, these correspondences can be found using a radius-based search to find neighboring points in 3D space. Another possible strategy is to generate these correspondences by reprojecting 3D points from one depth map to their neighboring views to re-establish the correspondences. These strategies generally perform similarly with only minor differences (an analysis of these approaches can be found in Section IV.D). Once we obtain the set of corresponding points as a point group, we establish a Markov Netlet, taking each point in the group as a node of the Netlet. Each node of this Netlet may be assigned a label representing an optimal view it should be associated with, such that points associated with the same view can be collapsed into a single point to remove the redundancies. It is expected that this labeling process should be smooth to ensure that the selected views are consistent and minimized. The edges of the Netlet can be weighted using varying factors such as color and distance. The simplest approach is based upon the spatial proximity of these points, and this will ensure the selected views are consistent when points are close to each other.

The input to our algorithm is a set of depth maps $\boldsymbol{d}_i \in \mathcal{D}$, each reflects from a view with its pose $pos_i$ and the point correspondences as a set of point groups $\mathcal{G}$ with $m$ nodes. For each $\boldsymbol{d}_i$, we represent each point as a node in the network. Then we enumerate connections between points within this $\mathcal{G}$, thus it creates a maximum of $C_2^m$ possible edges, representing the spatial relationship between points in 3D space. In particular, we hypothesize that these connected nodes are usually redundant and inconsistent since they are obtained from different depth maps. Thus, to ensure consistency, each node will have a number of candidate depth maps they can be associated with, and the idea is to assign this association to the same depth map, such that the position of this node (the 3D point) can be corrected to the corresponding 3D position from the same map as others in this point group $\mathcal{G}$. Thus, finding the associated view for each node becomes a labeling problem. For instance, if a node has three edges (with four potential depth maps it can associate with), it should be assigned to four potential labels. In the following, we describe the formulation of the proposed solution.

Let $X = \{\mathbf{p}_1, \mathbf{p}_2, \dots, \mathbf{p}_n\} \in \mathbb{R}^{3 \times n}$ be the set of 3D points, each corresponding to a depth value from their original depth images. Consider $Y = \{y_1, y_2, \dots, y_n\}$ as the point labels which represent a set of depth maps (views) $\mathcal{D} = \{\boldsymbol{d}_1, \dots, \boldsymbol{d}_k\}$ in the dataset, while $k$ is the number of views in the dataset. In particular, we regard each view as a label in our formulation, where we seek to infer a label for each point in the set $X$. Thus, the labels $Y$ can be inferred from the observed points $X$ by maximizing the posterior (MAP) probability to find the optimal estimation of $\hat{Y}$ as follows:

$$\hat{Y} = \arg\max_Y P(Y|X) \tag{1}$$

In a Bayesian framework, the equation above is equivalent to minimizing the energy function $E(Y|X)$ for a pairwise Markov Random Field (MRF) [31], which is represented as:

$$E(Y|X) = \sum_{i \in n} \varphi_i(y_i, \mathbf{p}_i) + \sum_{i \in n, j \in g_i} \varphi_{ij}(y_i, y_j) \tag{2}$$

where $\varphi_i(y_i, \mathbf{p}_i)$ denotes the unary term, $\varphi_{ij}(y_i, y_j)$ is the pairwise potential term, and $g_i$ is the point group containing the neighboring points of the point $\mathbf{p}_i$. The unary term defines an equal contribution of label candidate, which set each element of the summation as a value of 1, and this will allow the smooth/pairwise term to play the major role in deciding the final labels. The pairwise term integrates the spatial information into the Markov Netlets by penalizing inconsistent values between two connected points (i.e., two connected points in a local point group). The simplest formulation is to define this consistency as the Euclidean distance of these two points $(\mathbf{p}_i, \mathbf{p}_j)$ in 3D, such that close-by points are likely to be associated with the same label under a Gaussian weight, while far-apart points (bigger than $\tau$) do not need to comply with the smoothness (as shown in (3)):

$$\varphi_{ij}(y_i, y_j) = \begin{cases} 0, & \text{if } i = j \text{ or } \|\mathbf{p}_i - \mathbf{p}_j\|_2 > \tau \\ e^{-\|\mathbf{p}_i - \mathbf{p}_j\|_2}, & \text{otherwise} \end{cases} \quad (3)$$

where $\tau$ is a threshold, which we empirically set to 2 with the actual unit in the 3D space (i.e., meters). The cost function is minimized using the MRF parallel solver introduced in [32] which was found to be extremely efficient. After the optimization, a label is selected for each node in a Markov Netlet. Then, we apply a Netlet collapse process, which combines nodes belonging to the same label within a Netlet into one node as shown in Fig. 2.

---

**Algorithm 1.** Formulating Markov Netlets for depth fusion.

---

**Input:** $\mathcal{D}$ – set of depth maps;
    $\mathcal{P}$ – set of poses;
    $\mathcal{G}$ – set of point groups;
    $\mathcal{S}$ – set of superpixel clusters for each image
**Output:** $\mathcal{M}$ – Markov Netlet
**for** $d_i \in \mathcal{D},\ pos_i \in \mathcal{P},\ s_i \in \mathcal{S},\ g_i \in \mathcal{G}$ **do**
    $C_i \leftarrow GetCentroids(s_i)$
    **for** $\mathbf{c}_i \in C_i$ **do**
        $\mathbf{p}_i \leftarrow Get3DPoint(d_i, pos_i, \mathbf{c}_i)$
        $\mathcal{M} \leftarrow AddNode(\mathbf{p}_i)$
        $\mathcal{M} \leftarrow AddLabel(\mathbf{p}_i, l_i)$
        **for** $\mathbf{x} \in g_i$ **do**
            $\mathbf{x}_j \leftarrow GetCorrespondingPoint(g_i, \mathbf{c}_i)$
            $\mathbf{c}_j \leftarrow GetCentroid(s_j, \mathbf{x}_j)$
            $\mathbf{p}_j \leftarrow Get3DPoint(d_j, pos_j, \mathbf{c}_j)$
            $\mathcal{M} \leftarrow AddLabel(\mathbf{p}_j, l_j)$
            **if not** $Edge(\mathcal{G}, \mathbf{p}_i, \mathbf{p}_j)$ **and** $\|\mathbf{p}_i - \mathbf{p}_j\|_2 \le \tau$ **then**
                $w \leftarrow e^{-\|\mathbf{p}_i - \mathbf{p}_j\|_2}$
                $\mathcal{M} \leftarrow AddEdge(\mathbf{p}_i, \mathbf{p}_j, w)$

---

### C. Extension for Large-Scale Data Using Superpixel-Based Clustering

Despite that optimizing the Markov Netlets are extremely fast, while this is still computationally prohibitive when this is scaled to mapping projects involving high-resolution images with a large number of pixels (hundreds of millions). Thus, modeling each point in the depth map as a node can be computationally prohibitive. Therefore, reducing the number of nodes in the Netlets is necessary to achieve the scalability of the method. To achieve this goal, we utilize a clustering mechanism based on superpixels and formulate the Netlets using clusters as nodes. Utilizing superpixel clusters not only achieves scalability but also ensures the homogeneity of points inside each cluster. Precisely, we use the Simple Linear Iterative Clustering algorithm (SLIC) [33] to obtain homogenous local regions for each image in the dataset (i.e., superpixel clusters $\mathcal{S}$). SLIC algorithm utilizes the color information and pixel locations to create clusters with similar sizes. During the formulation of the Markov Netlets, instead of modeling each 3D point as a node, we regard each clustered region as the node to dramatically reduce the computation. In particular, we identify the centroid of each cluster to represent all neighboring points within the same cluster and then use the centroid as a node in the graph. Then, we build connectivity between 3D points following the same formulation explained in Section III.B as follows: for each cluster in a view/depth map $d_i$, we use its centroid $\mathbf{c}_i$ and find the corresponding points to $\mathbf{c}_i$ in neighboring views using the point group $g_i$. Then, we obtain the centroid $\mathbf{c}_j$ of neighboring points in $g_i$ to build an edge between both centroids $\mathbf{c}_i$ and $\mathbf{c}_j$. Note that the connectivity is built using the centroid of the cluster and the remaining points inside a cluster will not be used in the formulation of the Netlets. During the inference, all points within the same cluster are assigned to the same label as the one selected for the cluster's centroid. The detailed algorithm for formulating the Markov Netlets can be found in Algorithm 1, followed by the resolution of the MRF parallel solver [32] per Netlet.

## IV. Experimental Results

We evaluate our SAC approach on two public benchmark datasets against a few state-of-the-art depth fusion techniques. These two datasets include the ETH3D high-resolution multi-view benchmark [1] and the Dortmund aerial dataset [34]. For a fair comparison, these same depth maps will be used to produce fusion results by other methods throughout the experiments, and the accuracy against the LiDAR reference data is evaluated using a truncated F1-score (described in Section IV.A). The following methods are used in our evaluation:

- [15], a standard approach using geometric consistency verification: this approach reprojects points from each depth map $d_i$ to its neighboring views $d_j$, and verifies the consistency based on the following criteria: first, the projected depth $d_{ij}$ (projecting from points of $d_i$ to the view of $d_j$) should invalidate (thus remove) the corresponding depth value in neighboring view $d_j$, if $d_{ij}$ has a smaller value ($d_{ij}$ occludes it); second: if $d_{ij}$ is close enough to the corresponding depth in $d_j$, the corresponding depth value in $d_j$ should be removed due to redundancy, in which we set the threshold of 1% relative error (the readers are encouraged to review Fig. 4 in [15]).

- [13], a variant of geometric consistency verification operated in the disparity and normal space: this method reprojects points from each depth map $d_i$ to its neighboring views $d_j$, resulting in a disparity value. A point is considered consistent if the difference between the estimated disparity from the reprojection and the one computed from the dense matching step is below a threshold. In addition, this method considers local information of the pixel using the surface normal and regards a point to be consistent if the normal at the point in a view and its corresponding points at neighboring views differ by at most 30º. Thus, if a point passed the previous two conditions (disparity and normal differences) in at least two views (complied views), it will be updated by taking the average of points from these compiled views. Otherwise, the point is removed.

TABLE I

QUANTITATIVE RESULTS ON ETH3D HIGH-RESOLUTION MVS BENCHMARK. BOLD DENOTES THE METHOD WITH THE BEST F1 SCORES (%) FOR TWO DISTANCE THRESHOLDS, 2 CM AND 5 CM SEPARATED BY "/".

| Scene | [15] | [13] | [7] | Multi-ray | Ours |
|---|---|---|---|---|---|
| courtyard | 63.38 / 78.79 | 57.11 / 72.09 | 59.87 / 76.56 | 57.12 / 74.13 | 66.03 / 80.23 |
| delivery_area | 72.43 / 83.90 | 54.91 / 69.21 | 68.72 / 81.26 | 65.45 / 78.13 | 72.60 / 84.06 |
| electro | 33.38 / 46.23 | 7.14 / 9.82 | 31.36 / 43.06 | 13.49 / 19.64 | 32.97 / 46.00 |
| facade | 56.81 / 74.01 | 50.06 / 61.82 | 51.99 / 69.81 | 49.98 / 65.87 | 57.37 / 74.18 |
| kicker | 29.23 / 39.02 | 9.49 / 13.99 | 27.52 / 36.79 | 18.19 / 24.53 | 29.18 / 39.00 |
| meadow | 44.74 / 53.35 | 11.32 / 24.00 | 42.73 / 52.19 | 38.79 / 48.10 | 44.57 / 53.34 |
| office | 44.22 / 55.75 | 6.18 / 12.23 | 42.25 / 53.46 | 31.22 / 40.80 | 44.14 / 55.73 |
| pipes | 32.09 / 38.11 | 17.80 / 24.93 | 31.13 / 36.98 | 26.56 / 32.30 | 31.99 / 38.11 |
| playground | 36.63 / 44.04 | 9.62 / 14.62 | 34.55 / 42.77 | 27.21 / 32.80 | 36.50 / 44.02 |
| relief | 72.43 / 82.20 | 54.32 / 63.98 | 55.20 / 64.67 | 61.31 / 74.61 | 71.87 / 82.03 |
| relief_2 | 74.07 / 83.15 | 56.74 / 67.73 | 70.78 / 80.56 | 66.54 / 77.51 | 73.58 / 83.07 |
| terrace | 77.36 / 87.98 | 40.76 / 52.40 | 74.33 / 84.18 | 69.82 / 81.42 | 76.75 / 87.60 |
| terrains | 76.98 / 86.24 | 60.04 / 70.14 | 74.04 / 83.95 | 70.29 / 80.88 | 77.05 / 86.28 |
| Average | 54.90 / 65.60 | 33.50 / 42.84 | 51.11 / 62.02 | 45.84 / 56.21 | **54.97** / **65.67** |

- [7], a typical approach performs volumetric surface reconstruction based on TSDF: this approach operates the fusion in the point cloud space, and we converted individual depth maps into point clouds for TSDF fusion, and directly retrieve the fused point clouds.
- Multi-ray triangulation: this is an intuitive approach for image fusion that directly merges points from different views through ray triangulation. We implemented this approach based on the generated Netlets (described in Section III.B), which considers all points are multi-ray correspondences from different views, therefore can be directly triangulated to a single 3D point using the camera parameters and poses of these views [35].

### A. Evaluation Protocol

Assuming the availability of LiDAR reference data, we follow the evaluation protocol as described in [1], which measures the accuracy and completeness of the reconstructed point cloud using the F1 score as a single measure. The F1 score here is adapted as the harmonic mean $\frac{2(ac)}{(a+c)}$ where $a$ and $c$ denote the accuracy and completeness of the reconstruction respectively. The accuracy is measured as the fraction of reconstruction points whose distances against the LiDAR reference data are below a distance threshold, as explained in [1]. Therefore, a distance threshold classifies the reconstructed points as true positive (TP) (with distance smaller than the threshold) and false positive (FP) (with distance bigger than the threshold), thus to be used to compute the accuracy (TP in reconstructed points / # reconstructed points). The completeness computes the fraction of points in the LiDAR reference data, whose distance to the reconstructed points is below the given distance threshold, thus computed as (TP in reference points / # reference points). In our evaluation, we used 2cm and 5cm as the distance threshold for the ETH3D dataset and 1m and 2m for the Dortmund dataset, selected based on the resolution of the datasets. The readers are encouraged to read the evaluation protocol in [1] for more details.

### B. Results on ETH3D Dataset

This dataset consists of 13 scenes captured in indoor and outdoor settings using a Nikon D3X DSLR camera. The ground truth was acquired using a Faro Focus X 330 laser scanner. We resize the image to $3000 \times 2000$ for all scenes. We follow the evaluation protocol described in Section IV.A to measure the F1 score. The statistics shown in Table I reveal that our approach outperforms all methods in the majority of both indoor and outdoor scenes provided in the benchmark. The results show that the SAC approach improves the F1 score by up to 0.13% compared to the best existing method and by up to 64.09% compared to other methods, and this is largely attributed to the increased consistency of point clouds coming from a single depth map (fundamentals depicted in Fig. 1). As explained in Section III.C, our approach uses superpixel clusters to enhance its scalability to large format images. Since we use the centroid of a cluster as nodes in the Netlets, there may be an inconsistency between connected centroids in different views. To measure this impact, we conducted an experiment in which we do not use the superpixel clustering technique, by representing each point as a single node in a Netlet. Using the clustering technique only slightly reduces the average F1 score of all scenes from 55.32% to 54.97% for a 2-cm threshold and from 65.77% to 65.67% for a 5-cm threshold.

To demonstrate the drivers of the performance of our method, we visualize the homogeneity of points coming from a single depth map by color-coding the before-fusion and after-fusion point clouds based on each point's association with depth maps (labels). Fig. 3 shows a sample of visual results for two scenes from the ETH3D dataset, where points with the same color indicate these points come from the same depth map. We observe that the generated point clouds using our SAC fusion approach reflect more spatially coherent patterns in terms of their color labels (associated depth maps), while the before-fusion point clouds show a more random pattern reflecting that points are uniformly sampled from each depth map. This fact is further supported by the histogram in Fig. 4, which shows the number of samples with respect to which depth map they are

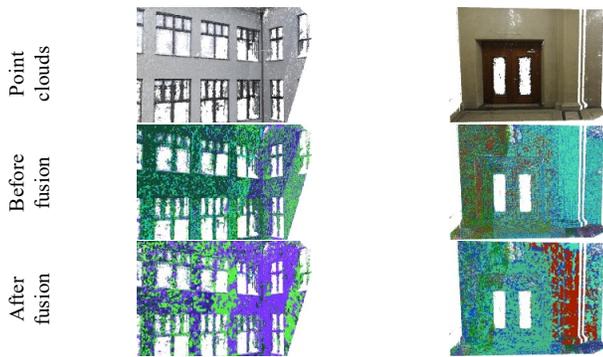

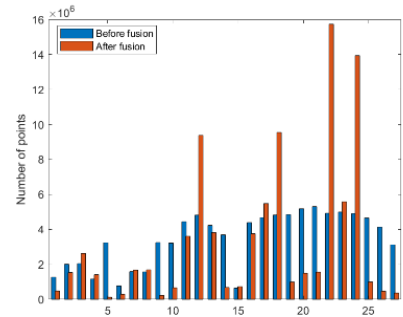

**Fig. 3.** Sample of visual results for the courtyard scene (first column) and the relief scene (second column) from the ETH3D dataset. The first row shows a sample of the generated point clouds, while the second and third rows show the corresponding points color-coded by different labels (selected images) before and after our SAC method. We observe that the generated points are more spatially coherent and consistent compared to the ones generated before the fusion step.

**Fig. 4.** The number of points generated per depth map before and after depth fusion using our SAC approach for the relief scene of the ETH3D dataset. By observing the change in the distribution of the number of points among different depth maps, we notice that after the fusion 4 views contain the majority of the generated points, compared to before fusion where most depth maps contain a relatively similar number of points.

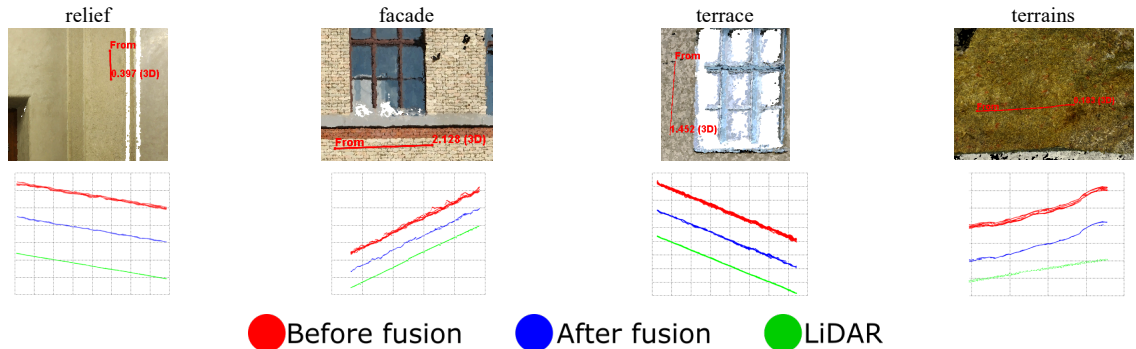

Before fusion    After fusion    LiDAR

**Fig. 5.** Illustration of the quality of generated point cloud before and after applying our SAC approach for different scenes of the ETH3D dataset. The second row shows profiles of the point clouds corresponding to the cross-section red segments shown in the first row (profiles of different sources are separated for better visualization).

derived from. It shows that in the after-fusion point clouds, the majority of points are derived from a limited number of depth maps (4-5 out of 27 in this particular example), while the before-fusion uniformly utilizes almost all the depth maps to produce the resulting point clouds. Therefore, the SAC algorithm achieves our expectation by attempting to select a reduced number of depth maps deriving the point clouds, thus ensuring the spatial consistency of final point clouds to yield higher accuracy following the rationale concluded from Fig. 1 (and associated texts). Fig. 5 visualizes cross-sections of several sample scenes before and after SAC fusion, we observe that the after-fusion results show much thinner and more consistent object structures, as opposed to those of before fusion, where object structures are presented by entangled points from various inconsistent depth maps.

### C. Results on Dortmund Dataset Airborne Images

Our proposed approach was also evaluated on a large-scene dataset from airborne sources (Dortmund dataset [34]). The data were acquired using a multi-camera system in an airborne platform and LiDAR data is used as the reference, covering an area of 1 km². An overview and the specifications of the dataset are shown and listed in Table II. Prior to computing the F1 score, we register the generated point clouds using different approaches to the LiDAR data using the robust iterative closest point method [36] to eliminate systematic misalignment. Table III shows the quantitative results for the entire area, following the same evaluation protocol described in Section IV.A. The results show that our SAC approach produces the best scores compared to other fusion methods with an improvement of up to 17.26% in the F1 score, further indicating the effectiveness of this method for airborne assets. Compared to the best existing method, our approach provides an increase of 2.07% in the F1 score. In addition, similar to Fig. 5, Fig. 6 visualizes the cross-sections of the before-fusion and after-fusion results on selected regions such as rooftop and ground regions, all showing that the after-fusion demonstrated clearer geometric structures. As compared to the LiDAR reference data, the mean-absolute error has improved from 0.46m to 0.40m (equivalently 12.09% of improvement). Further, we observe that the fusion has also effectively reduced about 18.27% of redundant points in this experiment, resulting in point cloud files consuming smaller storage.

## TABLE II
### DORTMUND AIRBORNE DATASET (SELECTED SUB-REGION) AND SPECIFICATIONS.

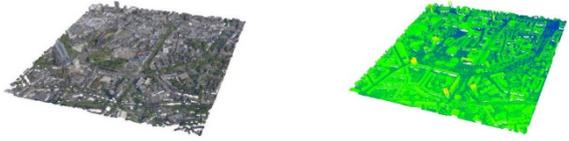

(a) Photogrammetric point cloud     (b) LiDAR point cloud

| Number of images | 16 nadir (N), 43 oblique (O) |
|---|---|
| Camera | Pentacam IGI |
| Image resolution | 6132 × 8176 pixels (N), 8176 ×6132 pixels(O) |
| Focal length | 50 mm (N), 80 mm (O) |
| Pixel size | 6 μm (N), 6 μm (O) |
| Overlap | 75/80 |
| Average GSD | 10 cm (N), 8-12 cm (O) |
| Reference data | Airborne LiDAR (10 pts/m²) |
| Area of coverage | ~1 km² |

(c) Specifications of the dataset. N refers to nadir camera, O refers to oblique camera. GSD: Ground sampling distance.

## TABLE III
### RESULTS ON DORTMUND DATASET. BOLD DENOTES THE METHOD WITH THE BEST F1 SCORES (%) FOR TWO DISTANCE THRESHOLDS, 1 M AND 2 M SEPARATED BY "/".

| [15] | [13] | [7] | Ours |
|---|---|---|---|
| 72.97/82.47 | 67.25/81.02 | 63.53/73.82 | **74.50/83.75** |

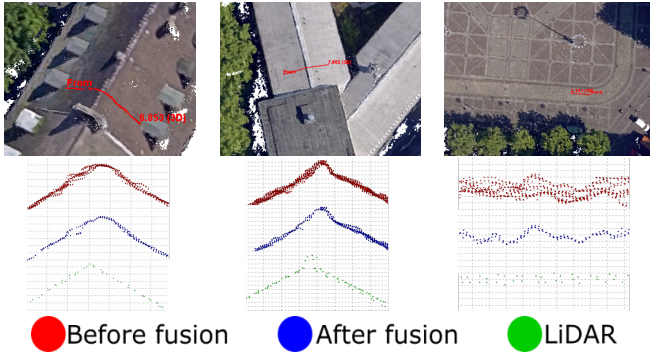

🔴 Before fusion    🔵 After fusion    🟢 LiDAR

**Fig. 6.** Illustration of the quality of generated point cloud before and after applying our SAC approach for different scenes of the Dortmund dataset. The second row shows profiles of the point clouds corresponding to the cross-section red segments shown in the first row (profiles of different sources are separated for better visualization).

### D. Strategies to Build Correspondences for Netlets

As mentioned in Section III.B, constructing the point group for Netlets requires building correspondences by traversing points across different depth maps, where different strategies can be used. In this experiment, we evaluate three strategies: the first strategy uses the dense correspondences estimated using the dense image matching step (SGM in our case) to reference identical points in space. The second strategy uses proximities of points from different depth maps in the object space. The third strategy builds the correspondence between points in 3D, but from the depth (view) space, which projects points from a depth map across different views to evaluate the difference, and this approach is similar to the geometric consistency verification approach [15]. The first strategy is straightforward to implement but requires lower-level access to the dense matching algorithms to store the dense correspondences across different images. For the second strategy, we construct an octree of all the point clouds and correspond points with a 3D distance lower than a threshold (10 cm in the experiments). The third strategy operates the correspondence search in the view space. We run our SAC algorithm with these different strategies and compare the results against the reference LiDAR by computing the cloud-to-cloud distance. Table IV shows the results, and we observe that all three strategies produce similar results with only marginal differences, thus indicating that the proposed method is robust towards approaches used to construct 3D point correspondences.

## TABLE IV
### ANALYZING DIFFERENT STRATEGIES TO BUILD EDGES IN THE NETWORK. WE MEASURE THE CLOUD-TO-CLOUD DISTANCE (M) BETWEEN DIFFERENT STRATEGIES.

| Building 3D point correspondences for Netlet construction | Dense matching (Strategy 1) | 3D proximity (Strategy 2) | view-level 3D proximity (Strategy 3) |
|---|---|---|---|
| Cloud-to-cloud distance | 0.2738 | 0.2791 | 0.2730 |

## V. CONCLUSION

This paper presents SAC (Select-and-Combine), a novel approach for depth map fusion based on Markov networks. In contrast to existing methods that seek arithmetic fusions, SAC develops a new paradigm that advocates the importance of selecting the most appropriate but minimally diverse depth map to be combined in the final point clouds. To do so, our approach builds adaptive connectivity between depth maps generated from different views in a form of a Markov network, called Markov Netlet because of its small dimension. In addition, we present an extended approach to scale this method for large datasets through superpixel clusters to improve efficiency. Because this approach regards fusion as a selection problem, it effectively combats noises following non-Gaussian distributions, at the same time minimizes the contributing depth maps to the final point clouds to retain spatial coherency.

We have demonstrated the superiority of our approach on both close-range (ETH3D dataset) and airborne datasets (Dortmund dataset). In the experiment using the ETH3D dataset, we showed that our approach outperforms the best existing method by up to 0.13% and other approaches by up to 64.09% in terms of accuracy and completeness measured using the F1 score. We demonstrated that our approach is scalable large-scale airborne assets in the Dortmund dataset experiment, in which we computed the results over a region of 1 km² and showed that the proposed method improved the F1-score by up to 2.07% relative to the best existing method. In both experiments, fusion results from our approach present the best F1-score over a few other comparing approaches. In addition, we have studied the influence of different strategies for Netlet construction and showed that our approach is insensitive to these variants. Our approach can be more effective when dealing with problems that have relatively sparse views in comparison to dense-view scenarios in SLAM or full-motion aerial videos. However, since our method optimizes local Markov Netlets, there are still opportunities to impose more global constraints to achieve improved depth consistency. Therefore, our future work will attempt to integrate more global formulations for depth selection.

## ACKNOWLEDGMENTS

The study was partially supported by an ONR grant (Award No. N000142012141 & N000142312670). The authors would like to acknowledge the provision of the Dortmund dataset by ISPRS and EuroSDR.